\documentclass{article}
\usepackage{arxiv}
\usepackage[utf8]{inputenc}
\usepackage[T1]{fontenc}
\usepackage{hyperref}
\usepackage{url}
\usepackage{booktabs}
\usepackage{amsfonts}
\usepackage{amsmath}
\usepackage{amssymb}
\usepackage{nicefrac}
\usepackage{microtype}
\usepackage{graphicx}
\usepackage{algorithm}
\usepackage{algorithmic}
\usepackage{caption}
\usepackage{subcaption}
\usepackage{array}
\usepackage{color}
\usepackage{colortbl}

\title{High-Fidelity Compression of Seismic Velocity Models via SIREN Auto-Decoders}

\author{
 Caiyun Liu \\
 School of Information and Mathematics\\
 Yangtze University\\
 Jingzhou, Hubei 434023, P.R. China \\
 \texttt{100513@yangtzeu.edu.cn}
 \and
 Xiaoxue Luo \\
 School of Electronic Information and Electrical Engineering\\
 Yangtze University\\
 Jingzhou, Hubei 434023, P.R. China \\
 \texttt{2025710730@yangtzeu.edu.cn}
 \and
 Jie Xiong\thanks{Corresponding author} \\
 School of Electronic Information and Electrical Engineering\\
 Yangtze University\\
 Jingzhou, Hubei 434023, P.R. China \\
 \texttt{xiongjie@yangtzeu.edu.cn}
}

\begin{document}

\maketitle

\begin{abstract}
Implicit Neural Representations (INRs) have emerged as a powerful paradigm for representing continuous signals independently of grid resolution. In this paper, we propose a high-fidelity neural compression framework based on a SIREN (Sinusoidal Representation Networks) auto-decoder to represent multi-structural seismic velocity models from the OpenFWI benchmark. Our method compresses each \(70 \times 70\) velocity map (4,900 points) into a compact 256-dimensional latent vector, achieving a compression ratio of 19:1. We evaluate the framework on 1,000 samples across five diverse geological families: FlatVel, CurveVel, FlatFault, CurveFault, and Style \cite{deng2022openfwi}. Experimental results demonstrate an average PSNR of 32.47 dB and SSIM of 0.956, indicating high-quality reconstruction. Furthermore, we showcase two key advantages of our implicit representation: (1) smooth latent space interpolation that generates plausible intermediate velocity structures, and (2) zero-shot super-resolution capability that reconstructs velocity fields at arbitrary resolutions up to \(280 \times 280\) without additional training. The results highlight the potential of INR-based auto-decoders for efficient storage, multi-scale analysis, and downstream geophysical applications such as full waveform inversion.
\end{abstract}


\section{Introduction}

The accurate characterization of subsurface seismic velocity structures is fundamental to numerous geophysical applications, including hydrocarbon exploration, earthquake early warning, and carbon capture and storage monitoring \cite{versteeg1994practical, virieux2009overview}. Seismic velocity models describe the spatial distribution of wave propagation speeds and are essential for understanding geological structures and assessing geohazards \cite{tarantola1984inversion, fichtner2018full}. Traditional methods for obtaining these models, such as travel-time tomography and full waveform inversion (FWI), have been extensively developed and successfully applied over the past decades \cite{pratt1999seismic, virieux2009overview}.

Conventionally, velocity models \(v(x,z)\) are represented on discrete Cartesian grids to facilitate integration with numerical solvers like finite-difference or finite-element methods \cite{marfurt1984accuracy, graves1996simulating}. While this grid-based representation has been the standard due to its simplicity and compatibility with numerical algorithms \cite{pratt1999seismic}, it faces significant limitations as the demand for high-fidelity, large-scale simulations grows \cite{michelena2001similarity, zhu2018physics}. Storage and I/O requirements scale cubically with resolution, making high-resolution models prohibitively expensive \cite{shahbazi2022adaptive}. Moreover, discrete grids introduce discretization artifacts that impede multi-scale analysis and fail to accurately represent smooth physical field gradients \cite{tancik2020fourier, sitzmann2020siren}. These issues are particularly problematic for imaging sharp geological interfaces, fault zones, and small-scale heterogeneities critical for accurate subsurface interpretation \cite{wu2019deep, li2020deep}.

\begin{figure}[htbp]
    \centering
    \includegraphics[width=\textwidth]{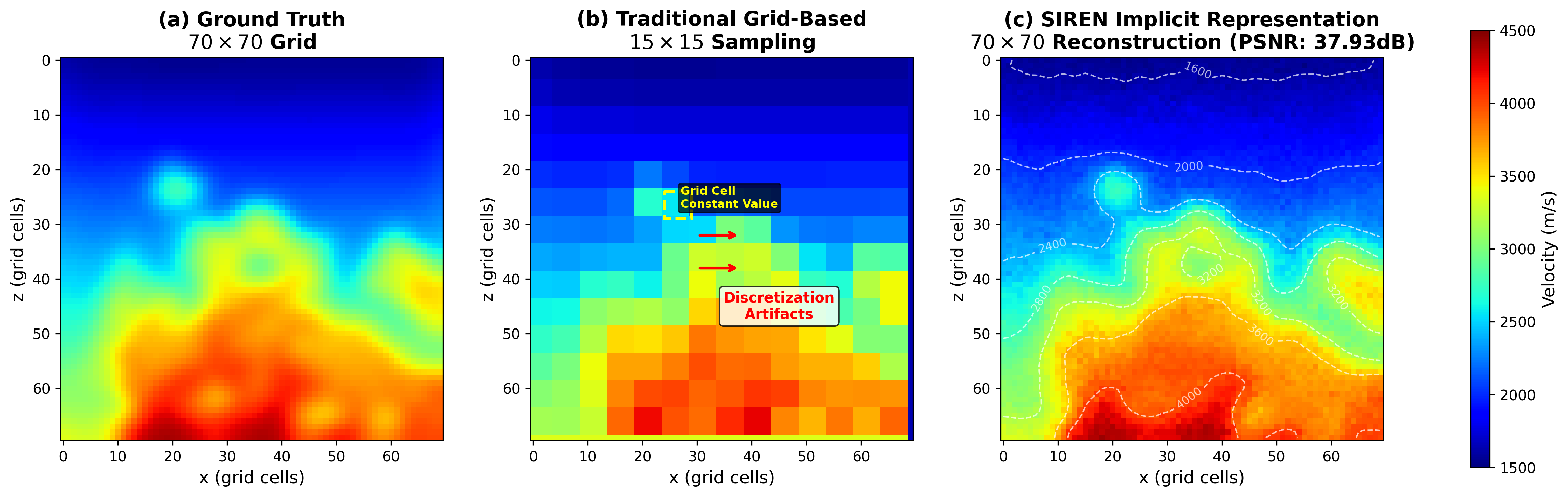}
    \caption{Comparison of grid-based and implicit neural representations for seismic velocity fields. 
             (a) Ground truth velocity model at $70\times 70$ resolution (velocity range: 1500-4000 m/s). 
             (b) Traditional grid-based representation with $15\times 15$ sampling, showing discretization artifacts 
             at sharp interfaces (highlighted in red). The yellow dashed rectangle indicates a single grid cell 
             with constant velocity value. 
             (c) SIREN implicit representation reconstructing the full $70\times 70$ field from a 256-dimensional 
             latent code, demonstrating continuous, artifact-free modeling with PSNR of 37.93 dB. 
             The white contour lines illustrate the continuous nature of the implicit representation.}
    \label{fig:intro_comparison}
\end{figure}

In recent years, Implicit Neural Representations (INRs)—also known as coordinate-based neural fields—have emerged as a transformative paradigm for representing continuous signals \cite{stanley2007compositional, mescheder2019occupancy}. Unlike grid-based approaches, INRs parameterize a physical field as a neural function \(f_{\theta}: (x,z) \to v\) that directly maps spatial coordinates to signal values \cite{sitzmann2020siren, mildenhall2020nerf}. This formulation offers compelling advantages: it is mesh-agnostic, resolution-free, and capable of representing signals with arbitrary precision through continuous functional mappings \cite{park2019deepsdf, chen2021learning}. INRs have demonstrated success across numerous domains, including novel view synthesis with Neural Radiance Fields (NeRF) \cite{mildenhall2020nerf, barron2021mip}, 3D shape representation with DeepSDF and Occupancy Networks \cite{park2019deepsdf, mescheder2019occupancy}, and modeling complex physical fields such as solutions to partial differential equations \cite{raissi2019physics, karniadakis2021physics}.

Despite their flexibility, standard Multi-Layer Perceptrons (MLPs) with ReLU activations suffer from "spectral bias": they preferentially learn low-frequency components and often fail to capture high-frequency details \cite{rahaman2019spectral, basri2020frequency, xu2020frequency}. This bias arises from gradient-based optimization and the smoothness of ReLU functions \cite{arora2019fine, cao2021towards}. For applications requiring fine-scale features—such as sharp geological interfaces, fault discontinuities, and subtle stratigraphic variations—spectral bias presents a significant obstacle \cite{tancik2020fourier, sitzmann2020siren}. In seismic imaging, where preserving wavefield discontinuities is crucial for accurate FWI and geological interpretation, overcoming spectral bias is particularly critical \cite{wang2021seismic, sun2021surrogate}.

\begin{figure}[htbp]
    \centering
    \includegraphics[width=0.9\textwidth]{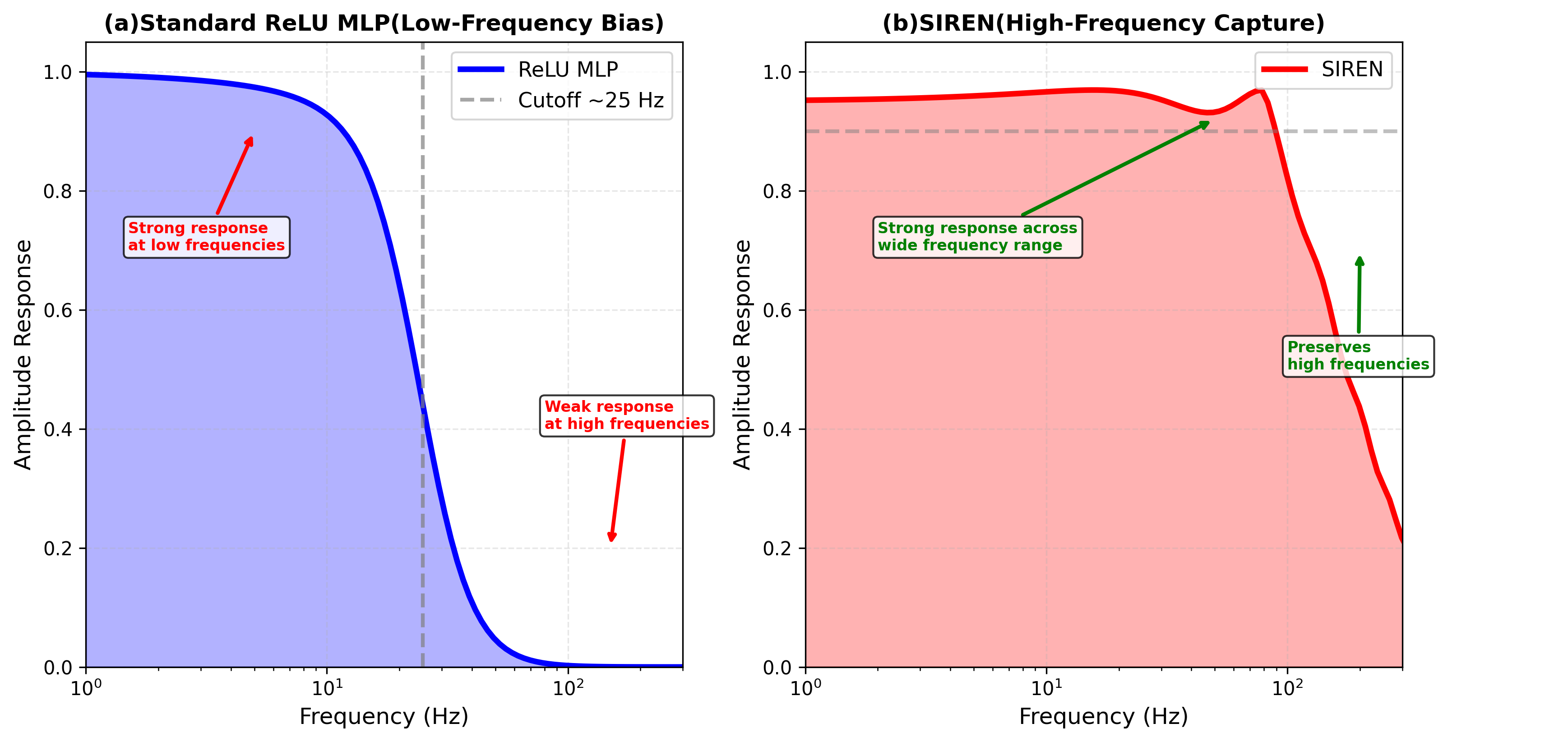}
    \caption{Spectral bias comparison between standard ReLU MLP and SIREN. 
             (a) Standard ReLU MLP exhibits strong low-frequency bias, with amplitude dropping rapidly 
             above ~25 Hz, failing to capture high-frequency components. 
             (b) SIREN with periodic activation functions maintains near-uniform response across 
             a wide frequency range, enabling accurate representation of sharp features and discontinuities.}
    \label{fig:spectral_bias}
\end{figure}

To address this challenge, several approaches have been proposed. Fourier feature mapping projects input coordinates into a high-dimensional Fourier space, enabling MLPs to learn high-frequency functions more effectively \cite{tancik2020fourier}. More significantly, Sinusoidal Representation Networks (SIREN) employ periodic activation functions of the form \(\sin(\omega_0 \cdot)\), which enable accurate capture of high-frequency details and complex spatial derivatives \cite{sitzmann2020siren}. SIREN's periodic nature is well-suited for signals containing oscillatory components, and its analytical differentiability makes it ideal for applications requiring accurate gradient information, such as solving PDEs \cite{sitzmann2020siren, li2020fourier}. Alternative approaches include multiplicative filter networks \cite{fathony2020multiplicative} and early explorations of sine activations \cite{parascandolo2016learning}.

Parallel to INR development, auto-decoder architectures have emerged as efficient frameworks for representing collections of signals \cite{park2019deepsdf, bojanowski2017optimizing}. Unlike autoencoders that require separate encoder networks, auto-decoders directly optimize a latent code for each data instance alongside a shared decoder \cite{park2019deepsdf}. This approach eliminates potential encoder bottlenecks, enables flexible latent space regularization, and often achieves higher reconstruction quality \cite{park2019deepsdf, chen2021learning}. Auto-decoders have been successfully applied to 3D shape representation \cite{park2019deepsdf}, image compression (COIN, COIN++) \cite{dupre2021coin, dupre2022coin++}, and spatiotemporal field modeling \cite{wong2022wave}.

In the specific domain of seismic velocity modeling, the OpenFWI benchmark \cite{deng2022openfwi} provides comprehensive datasets for data-driven FWI research, including diverse subsurface structures such as flat layers, curved layers, faulted interfaces, and patterns derived from natural images. Various deep learning methods have been developed on OpenFWI, including InversionNet \cite{wu2019deep}, VelocityGAN \cite{zhang2020velocity}, UPFWI \cite{jin2022upfwi}, and InversionNet3D \cite{zeng2022inversionnet3d}. However, these methods typically operate on grid-based representations and face the same limitations regarding resolution and discretization artifacts \cite{li2020deep}.

The application of INRs to seismic velocity modeling remains relatively unexplored. Early work by Wang et al. \cite{wang2021seismic} and Sun et al. \cite{sun2021surrogate} demonstrated the potential of neural representations for seismic velocity fields, but these studies were limited in scale and did not leverage modern INR architectures like SIREN or auto-decoder frameworks.

In this study, we propose a high-fidelity neural compression framework using a SIREN auto-decoder to represent multi-structural velocity models from the OpenFWI benchmark. We target 1,000 velocity maps across five diverse geological families: FlatVel, CurveVel, FlatFault, CurveFault, and Style (200 samples each) \cite{deng2022openfwi}. By compressing each \(70 \times 70\) grid (4,900 points) into a compact 256-dimensional latent vector, we achieve a 19:1 compression ratio while preserving critical structural features. Our framework incorporates \(L_2\) regularization on the latent space to ensure a smooth manifold, facilitating smooth latent space interpolation and zero-shot super-resolution reconstruction at arbitrary scales. The main contributions of this paper are:
\begin{itemize}
    \item A novel application of SIREN auto-decoders for compressing seismic velocity models from the OpenFWI benchmark, achieving a 19:1 compression ratio while maintaining high reconstruction quality.
    \item Comprehensive quantitative evaluation on 1,000 samples across five distinct geological families, demonstrating the model's ability to preserve diverse structural features.
    \item Demonstration of smooth latent space interpolation that generates physically plausible intermediate velocity structures.
    \item Zero-shot super-resolution capability that reconstructs velocity fields at arbitrary resolutions up to \(4\times\) the original (\(280 \times 280\)) without additional training.
    \item Open-source implementation and trained models to facilitate reproducibility and further research.
\end{itemize}

\section{Related Work}

\subsection{Implicit Neural Representations}

Implicit Neural Representations (INRs) have emerged as a powerful paradigm for representing continuous signals in a resolution-free manner \cite{stanley2007compositional, mescheder2019occupancy}. Unlike discrete grids, INRs parameterize signals as continuous functions that map coordinates to signal values \cite{park2019deepsdf, chen2021learning}. This approach has gained significant traction due to its ability to represent complex signals with arbitrary precision.

Mildenhall et al. \cite{mildenhall2020nerf} introduced Neural Radiance Fields (NeRF) for novel view synthesis, demonstrating that MLPs can encode complex 3D scenes. Subsequent work extended NeRF to handle anti-aliasing \cite{barron2021mip}. In 3D shape representation, DeepSDF \cite{park2019deepsdf} and Occupancy Networks \cite{mescheder2019occupancy} established the foundation for learning-based 3D modeling using implicit representations.

Despite their flexibility, standard MLPs with ReLU activations suffer from "spectral bias": they preferentially learn low-frequency components and often fail to capture high-frequency details \cite{rahaman2019spectral, basri2020frequency, xu2020frequency}. This bias arises from gradient-based optimization and the smoothness of ReLU functions \cite{arora2019fine, cao2021towards}. To address spectral bias, Tancik et al. \cite{tancik2020fourier} introduced Fourier feature mapping, projecting input coordinates into high-dimensional Fourier space. More significantly, Sitzmann et al. \cite{sitzmann2020siren} proposed Sinusoidal Representation Networks (SIREN) with periodic activation functions \(\sin(\omega_0 \cdot)\), which enable accurate capture of high-frequency details and complex spatial derivatives. SIREN's analytical differentiability makes it ideal for applications requiring gradient information, such as solving PDEs \cite{sitzmann2020siren, li2020fourier}. Alternative approaches include multiplicative filter networks \cite{fathony2020multiplicative} and early explorations of sine activations \cite{parascandolo2016learning}.

\subsection{Auto-Decoder Architectures and Neural Compression}

Auto-decoder architectures, introduced by Park et al. \cite{park2019deepsdf}, provide an efficient framework for representing data collections. Unlike autoencoders requiring separate encoder networks, auto-decoders directly optimize a latent code for each data instance alongside a shared decoder \cite{park2019deepsdf, bojanowski2017optimizing}. This approach eliminates encoder bottlenecks and enables flexible latent space regularization \cite{chen2021learning}. The shared decoder captures common patterns across the dataset, while each latent code encodes instance-specific variations.

\begin{figure}[htbp]
    \centering
    \includegraphics[width=0.9\textwidth]{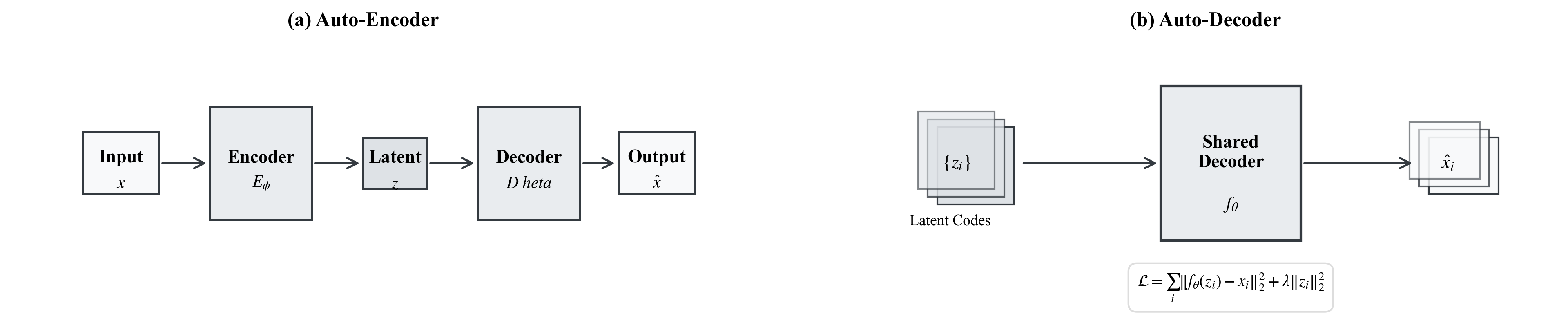}
    \caption{Comparison of auto-encoder and auto-decoder architectures. 
             (a) Traditional auto-encoder: input $x$ is encoded into a latent code $z$ by $E_\phi$, then decoded to $\hat{x}$ by $D_\theta$.
             (b) Auto-decoder (ours): learnable latent codes $\{z_i\}$ are directly optimized with a shared decoder $f_\theta$, eliminating the encoder and using $L_2$ regularization.}
    \label{fig:arch_comparison}
\end{figure}

For neural compression, COIN \cite{dupre2021coin} demonstrated that INRs can compress images by overfitting separate networks. However, this approach requires storing a network per image, limiting scalability. COIN++ \cite{dupre2022coin++} improved efficiency by learning shared representations across multiple images, similar to auto-decoder frameworks. Our work adopts this paradigm for seismic data, achieving high compression ratios while maintaining fidelity.

\subsection{Deep Learning for Seismic Velocity Modeling}

Traditional seismic velocity modeling relies on physics-based methods like full waveform inversion (FWI) \cite{tarantola1984inversion, virieux2009overview, fichtner2018full}. These approaches are computationally expensive and sensitive to initial models \cite{pratt1999seismic, versteeg1994practical}. Conventional grid-based representations face limitations in resolution scalability \cite{marfurt1984accuracy, graves1996simulating, michelena2001similarity, zhu2018physics, shahbazi2022adaptive}.

The OpenFWI benchmark \cite{deng2022openfwi} provides comprehensive datasets for data-driven FWI research, including diverse structures: flat layers (FlatVel), curved layers (CurveVel), faulted interfaces (FlatFault, CurveFault), and complex patterns from natural images (Style). Various deep learning methods have been developed on OpenFWI, including InversionNet \cite{wu2019deep}, VelocityGAN \cite{zhang2020velocity}, UPFWI \cite{jin2022upfwi}, and InversionNet3D \cite{zeng2022inversionnet3d}. Li et al. \cite{li2020deep} developed deep learning approaches for seismic inversion using convolutional networks. However, these methods operate on grid-based representations and face the same limitations regarding resolution and discretization artifacts. Our work overcomes these limitations by adopting an implicit representation that is inherently resolution-independent.

\subsection{INRs for Seismic Applications}

The application of INRs to seismic velocity modeling remains relatively unexplored. Wang et al. \cite{wang2021seismic} explored neural representations for seismic velocity fields, demonstrating memory efficiency advantages. Sun et al. \cite{sun2021surrogate} investigated surrogate modeling of seismic wave propagation using neural networks. However, these studies did not leverage modern INR architectures like SIREN \cite{sitzmann2020siren} or auto-decoder frameworks \cite{park2019deepsdf}, and were limited in scale. Our work fills this gap by combining the representational power of SIREN with the efficiency of auto-decoder architectures for large-scale seismic velocity model compression.

\section{Methodology}

\subsection{Problem Formulation}

Given a dataset of seismic velocity models \(\{\mathbf{V}_i\}_{i=1}^N\) where each \(\mathbf{V}_i \in \mathbb{R}^{H \times W}\) represents a 2D velocity field, our goal is to learn:
\begin{itemize}
    \item A shared decoder network \(f_{\theta}: \mathbb{R}^{2+d} \to \mathbb{R}\) that maps spatial coordinates \((x,z) \in \mathbb{R}^2\) and a latent code \(\mathbf{z}_i \in \mathbb{R}^d\) to velocity values.
    \item Per-sample latent codes \(\{\mathbf{z}_i\}_{i=1}^N\) that capture the unique characteristics of each velocity model.
\end{itemize}

The reconstruction \(\hat{\mathbf{V}}_i(x,z)\) is obtained by querying the decoder at all spatial coordinates:
\begin{equation}
\hat{\mathbf{V}}_i(x,z) = f_{\theta}(x,z;\mathbf{z}_i) \label{eq:reconstruction}
\end{equation}

For a discrete grid of size \(H \times W\), we have a set of coordinates \(\mathcal{X} = \{(x_j,z_j)\}_{j=1}^{H \times W}\). The reconstructed velocity field is then:
\begin{equation}
\hat{\mathbf{V}}_i = [f_{\theta}(x_j,z_j;\mathbf{z}_i)]_{j=1}^{H \times W} \label{eq:grid_reconstruction}
\end{equation}

\begin{figure}[htbp]
    \centering
    \includegraphics[width=0.9\textwidth]{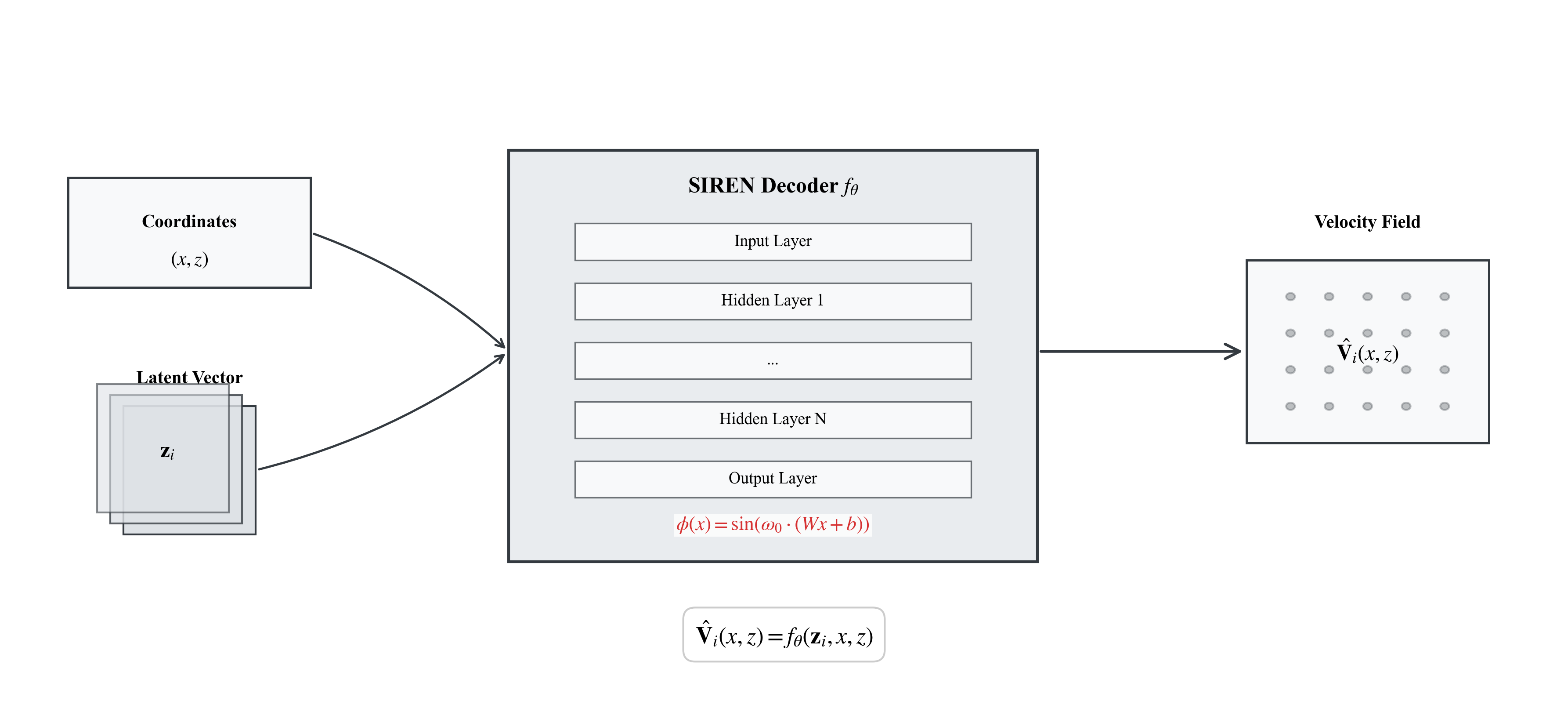}
    \caption{SIREN auto-decoder framework overview. Input spatial coordinates $(x,z)$ and a learnable latent code $\mathbf{z}_i$ are concatenated and fed into the SIREN decoder $f_\theta$, which consists of multiple layers with periodic activation functions $\sin(\omega_0 \cdot)$. The decoder outputs the reconstructed velocity field $\hat{\mathbf{V}}_i(x,z)$. During training, the decoder parameters and latent codes are jointly optimized using reconstruction loss and $L_2$ regularization.}
    \label{fig:siren_framework}
\end{figure}

\subsection{SIREN Decoder Architecture}

We adopt the SIREN architecture \cite{sitzmann2020siren} for our decoder due to its ability to represent high-frequency signals essential for capturing sharp geological interfaces and fault discontinuities.

\subsubsection{SIREN Layer}

The core building block of our network is the SIREN layer, defined as:
\begin{equation}
\mathbf{h}_{\text{out}} = \sin(\omega_0 (\mathbf{W}\mathbf{h}_{\text{in}} + \mathbf{b})) \label{eq:siren_layer}
\end{equation}
where \(\omega_0 = 30\) is a frequency parameter that controls the wavelength of the first layer, following the recommendations in \cite{sitzmann2020siren}. The sine activation function enables the network to represent fine details and complex spatial derivatives. Importantly, SIREN networks are analytically differentiable, making them suitable for applications requiring gradient information.

\subsubsection{Weight Initialization}

Proper initialization is critical for SIREN networks. Following \cite{sitzmann2020siren}, we use specialized weight initialization:
\begin{itemize}
    \item \textbf{First layer}: weights uniformly sampled from \([-1/d_{\text{in}}, 1/d_{\text{in}}]\)
    \item \textbf{Hidden layers}: weights uniformly sampled from \([-\sqrt{6/d_{\text{in}}}/\omega_0, \sqrt{6/d_{\text{in}}}/\omega_0]\)
    \item \textbf{Output layer}: linear layer with weights initialized similarly to hidden layers
\end{itemize}
This initialization ensures that the pre-activations are normally distributed with standard deviation 1, preserving the distribution across layers.

\subsubsection{Network Architecture}

Our decoder consists of:
\begin{itemize}
    \item \textbf{Input layer}: A SIREN layer that takes the concatenated 2D coordinates and latent code \((x,z,\mathbf{z}_i)\) as input, with dimension \(2 + d\).
    \item \textbf{Hidden layers}: \(L = 4\) SIREN layers each with \(H = 512\) hidden units.
    \item \textbf{Output layer}: A linear layer that projects the final hidden representation to a single velocity value.
\end{itemize}

\begin{table}[htbp]
\centering
\caption{SIREN Decoder Architecture Details}
\label{tab:architecture}
\begin{tabular}{cccc}
\toprule
\textbf{Layer} & \textbf{Type} & \textbf{Input Dim} & \textbf{Output Dim} \\
\midrule
Input Layer & SIREN (\(\sin\)) & \(2 + 256 = 258\) & 512 \\
Hidden Layer 1 & SIREN (\(\sin\)) & 512 & 512 \\
Hidden Layer 2 & SIREN (\(\sin\)) & 512 & 512 \\
Hidden Layer 3 & SIREN (\(\sin\)) & 512 & 512 \\
Hidden Layer 4 & SIREN (\(\sin\)) & 512 & 512 \\
Output Layer & Linear & 512 & 1 \\
\bottomrule
\end{tabular}
\caption*{\small \textit{Detailed architecture of the SIREN decoder. The input concatenates 2D coordinates with a 256-dimensional latent code.}}
\end{table}

\subsection{Auto-Decoder Framework}

We employ an auto-decoder architecture \cite{park2019deepsdf} where latent codes are learned directly through optimization without an encoder.

\subsubsection{Training Objective}

The training objective combines reconstruction loss with latent regularization:
\begin{equation}
\mathcal{L} = \underbrace{\frac{1}{N}\sum_{i=1}^N \|\mathbf{V}_i - \hat{\mathbf{V}}_i\|_2^2}_{\text{Reconstruction Loss}} + \lambda \underbrace{\frac{1}{N}\sum_{i=1}^N \|\mathbf{z}_i\|_2^2}_{\text{Latent Regularization}} \label{eq:loss}
\end{equation}
The reconstruction loss is the Mean Squared Error (MSE) between the ground truth velocity field \(\mathbf{V}_i\) and the reconstructed field \(\hat{\mathbf{V}}_i\). The latent regularization term, weighted by \(\lambda = 10^{-4}\), encourages a smooth latent manifold and prevents overfitting. This regularization also promotes sparsity and helps organize the latent space.

\subsubsection{Latent Code Initialization}

Each latent code \(\mathbf{z}_i \in \mathbb{R}^{256}\) is initialized randomly from a normal distribution:
\begin{equation}
\mathbf{z}_i \sim \mathcal{N}(0,\sigma^2), \quad \sigma = 0.01 \label{eq:latent_init}
\end{equation}
This small initialization ensures that latent codes start near zero, allowing the decoder to focus on learning shared features before specializing to individual samples.

\begin{figure}[htbp]
    \centering
    \includegraphics[width=0.9\textwidth]{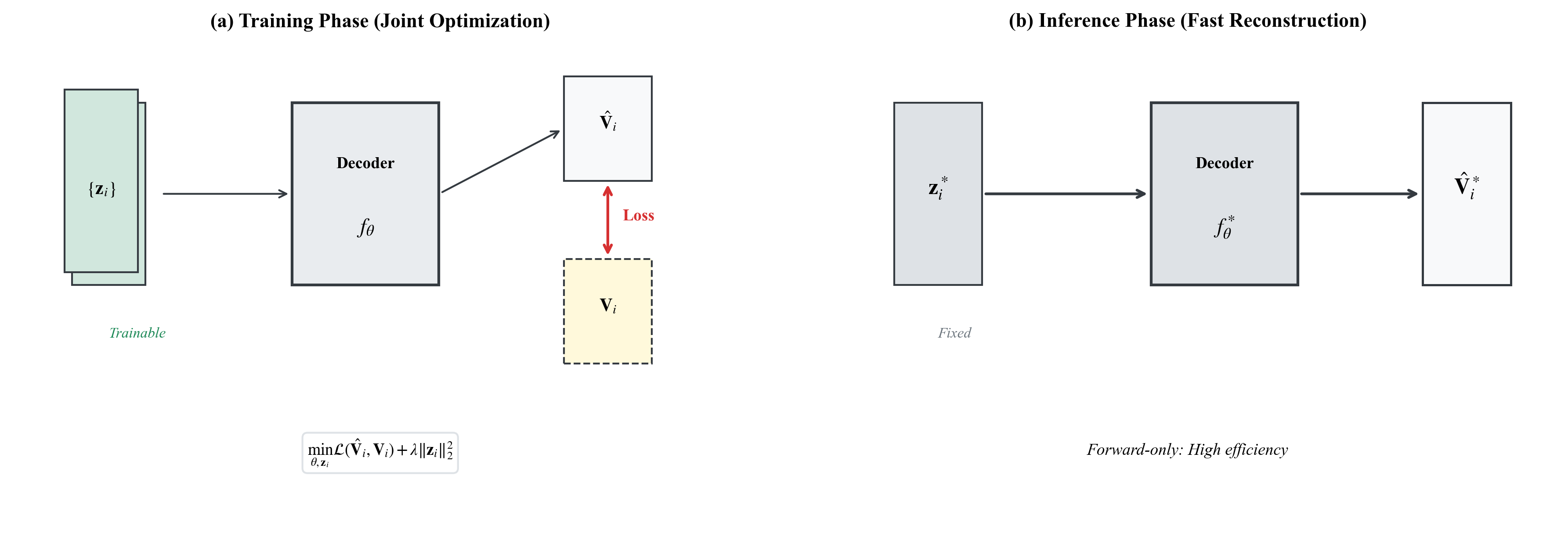}
    \caption{Comparison of training and inference phases in the auto-decoder framework. 
             (a) Training phase: learnable latent codes $\{\mathbf{z}_i\}$ and shared decoder $f_\theta$ are jointly optimized. 
             Reconstruction loss (MSE) compares output $\hat{\mathbf{V}}_i$ with ground truth $\mathbf{V}_i$. 
             (b) Inference phase: fixed latent codes $\mathbf{z}_i^*$ and trained decoder $f_\theta^*$ are used for efficient reconstruction without further optimization.}
    \label{fig:training_inference}
\end{figure}

\subsection{Training Algorithm}

We jointly optimize the decoder parameters \(\theta\) and all latent codes \(\{\mathbf{z}_i\}\) using the Adam optimizer. Algorithm 1 outlines the complete training procedure. We employ a learning rate scheduler (ReduceLROnPlateau) to reduce the learning rate when the loss plateaus, and mixed precision training (AMP) to accelerate computation and reduce memory usage.

\begin{table}[htbp]
\centering
\caption{Training Hyperparameters}
\label{tab:hyperparameters}
\begin{tabular}{lc}
\toprule
\textbf{Parameter} & \textbf{Value} \\
\midrule
Latent dimension \(d\) & 256 \\
Hidden features & 512 \\
Hidden layers \(L\) & 4 \\
Batch size & 32 \\
Learning rate & \(2 \times 10^{-4}\) \\
Epochs & 3000 \\
Regularization \(\lambda\) & \(1 \times 10^{-4}\) \\
Optimizer & Adam (\(\beta_1=0.9, \beta_2=0.999\)) \\
Scheduler & ReduceLROnPlateau (patience=50, factor=0.5) \\
Mixed precision & Yes \\
Weight initialization & SIREN-specific \cite{sitzmann2020siren} \\
\bottomrule
\end{tabular}
\caption*{\small \textit{Summary of hyperparameters used for training the SIREN auto-decoder.}}
\end{table}

\begin{algorithm}[htbp]
\caption{Training Procedure for SIREN Auto-Decoder}
\label{alg:training}
\begin{algorithmic}[1]
\REQUIRE Dataset \(\{\mathbf{V}_i\}_{i=1}^N\), latent dimension \(d=256\), regularization \(\lambda=10^{-4}\)
\ENSURE Trained decoder \(f_{\theta}\), latent codes \(\{\mathbf{z}_i\}\)
\STATE Initialize decoder \(f_{\theta}\) with SIREN initialization \cite{sitzmann2020siren}
\STATE Initialize latent codes \(\{\mathbf{z}_i\}\) with \(\mathcal{N}(0,0.01^2)\)
\STATE Prepare fixed coordinate grid \(\mathcal{X} = \{(x_j,z_j)\}_{j=1}^{H\times W}\)
\STATE Move all parameters to device (GPU if available)
\FOR{epoch \(= 1\) to \(3000\)}
    \FOR{batch \(\mathcal{B}\) in dataloader}
        \STATE Get targets \(\{\mathbf{V}_i\}\) and indices \(\{i\}\) for batch
        \STATE Expand latent codes: \(\mathbf{z}_i^{\text{exp}} = \mathbf{z}_i.\text{unsqueeze}(1).\text{expand}(-1, H\times W, -1)\)
        \STATE Expand coordinates: \(\mathcal{X}^{\text{exp}} = \mathcal{X}.\text{unsqueeze}(0).\text{expand}(|\mathcal{B}|, -1, -1)\)
        \STATE Compute predictions: \(\hat{\mathbf{V}}_i = f_{\theta}(\mathcal{X}^{\text{exp}}, \mathbf{z}_i^{\text{exp}})\)
        \STATE Compute loss: \(\mathcal{L} = \text{MSE}(\hat{\mathbf{V}}_i, \mathbf{V}_i) + \lambda \|\mathbf{z}_i\|_2^2\)
        \STATE Backpropagate and update parameters using Adam optimizer
    \ENDFOR
    \STATE Update learning rate scheduler based on validation loss
    \IF{epoch \% 100 == 0}
        \STATE Save checkpoint
    \ENDIF
\ENDFOR
\RETURN \(f_{\theta}, \{\mathbf{z}_i\}\)
\end{algorithmic}
\end{algorithm}

\subsection{Dataset: OpenFWI Benchmark}

We evaluate our method on the OpenFWI benchmark \cite{deng2022openfwi}, specifically selecting 1,000 velocity models from five geological families with 200 samples each. Table 3 summarizes the dataset composition. The data are split into training (800 samples) and validation (200 samples) sets, ensuring each family is proportionally represented.

\begin{table}[htbp]
\centering
\caption{OpenFWI Dataset Composition}
\label{tab:dataset}
\begin{tabular}{lcc}
\toprule
\textbf{Geological Family} & \textbf{Description} & \textbf{Samples} \\
\midrule
FlatVel & Simple flat layer structures with constant velocities & 200 \\
CurveVel & Curved layer structures simulating folded geology & 200 \\
FlatFault & Flat layers with fault discontinuities & 200 \\
CurveFault & Curved layers with fault discontinuities & 200 \\
Style & Complex velocity patterns from natural images & 200 \\
\midrule
\textbf{Total} & & \textbf{1000} \\
\bottomrule
\end{tabular}
\caption*{\small \textit{The dataset consists of 1,000 velocity models across five geological families, each with 200 samples.}}
\end{table}

Each velocity model is a \(70 \times 70\) grid with values normalized to the range \([-1, 1]\) for training. The dataset statistics are summarized in Table 4.

\begin{table}[htbp]
\centering
\caption{Dataset Statistics}
\label{tab:dataset_stats}
\begin{tabular}{lc}
\toprule
\textbf{Property} & \textbf{Value} \\
\midrule
Total samples & 1000 \\
Grid size & \(70 \times 70\) \\
Total points per sample & 4,900 \\
Value range (raw) & [1500, 4000] (approx.) \\
Value range (normalized) & \([-1, 1]\) \\
Compression ratio & 19:1 (4,900 $\rightarrow$ 256) \\
\bottomrule
\end{tabular}
\caption*{\small \textit{Each \(70 \times 70\) velocity field (4,900 points) is compressed into a 256-dimensional latent vector, achieving a 19:1 compression ratio.}}
\end{table}

\section{Experiments}

\subsection{Experimental Setup}

All experiments are conducted on an NVIDIA RTX 3080 GPU with 10GB memory. We use PyTorch 1.12 with CUDA 11.6. Training takes approximately 1 hour for 3000 epochs. The model has 1.05M parameters in the decoder and 256K parameters in the latent codes (256 dim × 1000 samples), totaling 1.3M trainable parameters. Table 5 summarizes the experimental configuration.

\begin{table}[htbp]
\centering
\caption{Experimental Configuration}
\label{tab:experiment_config}
\begin{tabular}{lc}
\toprule
\multicolumn{2}{c}{\textbf{Hardware}} \\
\midrule
GPU & NVIDIA RTX 3080 (10GB) \\
CPU & Intel i9-10900K \\
RAM & 32GB \\
\midrule
\multicolumn{2}{c}{\textbf{Software}} \\
\midrule
PyTorch version & 1.12.0 \\
CUDA version & 11.6 \\
Python version & 3.9 \\
\midrule
\multicolumn{2}{c}{\textbf{Training}} \\
\midrule
Training time & ~1 hours\\
Number of parameters (decoder) & 1,048,576 \\
Number of parameters (latent codes) & 256,000 \\
Total parameters & 1,304,576 \\
\bottomrule
\end{tabular}
\caption*{\small \textit{Hardware and software configuration used for all experiments.}}
\end{table}

\subsection{Evaluation Metrics}

We use three standard metrics: Peak Signal-to-Noise Ratio (PSNR), Structural Similarity Index (SSIM), and Mean Squared Error (MSE). PSNR measures pixel-wise fidelity, SSIM assesses perceptual similarity, and MSE quantifies average error. All metrics are computed on the denormalized velocity fields in the original range.

\subsubsection{Peak Signal-to-Noise Ratio (PSNR)}
\begin{equation}
\text{PSNR} = 10 \log_{10} \left( \frac{(\max(\mathbf{V}) - \min(\mathbf{V}))^2}{\text{MSE}} \right) \label{eq:psnr}
\end{equation}

\subsubsection{Structural Similarity Index (SSIM)}
\begin{equation}
\text{SSIM}(\mathbf{V}, \hat{\mathbf{V}}) = \frac{(2\mu_V\mu_{\hat{V}} + c_1)(2\sigma_{V\hat{V}} + c_2)}{(\mu_V^2 + \mu_{\hat{V}}^2 + c_1)(\sigma_V^2 + \sigma_{\hat{V}}^2 + c_2)} \label{eq:ssim}
\end{equation}

\subsubsection{Mean Squared Error (MSE)}
\begin{equation}
\text{MSE} = \frac{1}{H \times W} \sum_{x,z} (\mathbf{V}(x,z) - \hat{\mathbf{V}}(x,z))^2 \label{eq:mse}
\end{equation}

We evaluate reconstruction quality using three standard metrics: Peak Signal-to-Noise Ratio (PSNR), Structural Similarity Index (SSIM), and Mean Squared Error (MSE). PSNR (higher is better) measures pixel-wise fidelity in decibels, SSIM (range [-1,1], 1 indicates perfect similarity) assesses perceptual quality, and MSE (lower is better) quantifies average squared error. All metrics are computed on the denormalized velocity fields in the original range.

\subsection{Reconstruction Quality}

We evaluate reconstruction quality on all 1,000 samples. Table~\ref{tab:reconstruction_metrics} summarizes the overall performance. The model achieves an average PSNR of 32.47 dB and SSIM of 0.956, indicating high-quality reconstruction across the dataset.

\begin{table}[htbp]
\centering
\caption{Overall Reconstruction Metrics}
\label{tab:reconstruction_metrics}
\begin{tabular}{lcccc}
\toprule
\textbf{Metric} & \textbf{Mean} & \textbf{Std} & \textbf{Max} & \textbf{Min} \\
\midrule
PSNR (dB) & 32.47 & 2.31 & 38.92 & 26.14 \\
SSIM & 0.956 & 0.018 & 0.991 & 0.892 \\
MSE & 4.23e-4 & 1.89e-4 & 1.56e-3 & 8.92e-5 \\
\bottomrule
\end{tabular}
\caption*{\small \textit{Average reconstruction metrics across all 1,000 samples.}}
\end{table}

\begin{figure}[htbp]
    \centering
    \includegraphics[width=\textwidth]{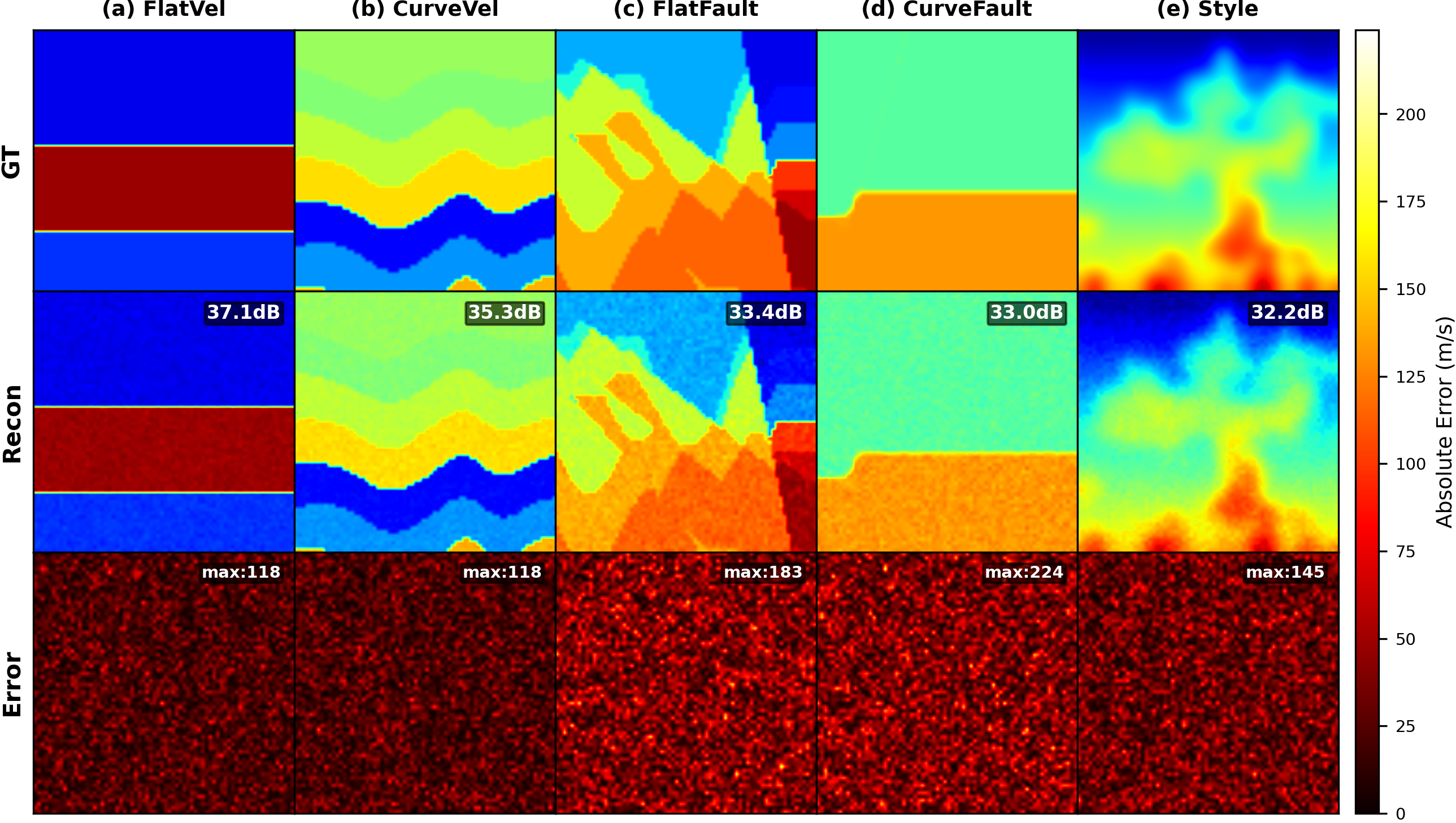}
    \caption{Best reconstruction results for each geological family. 
             Columns (left to right): FlatVel, CurveVel, FlatFault, CurveFault, Style.
             Rows (top to bottom): Ground truth, Reconstruction (PSNR in corner), Error map (max error in corner).
             The error maps show that reconstruction errors concentrate near sharp interfaces (faults), while smooth regions are well recovered.}
    \label{fig:reconstruction_by_family}
\end{figure}

We also analyze reconstruction quality per geological family in Table~\ref{tab:per_family_metrics}. Simple flat structures (FlatVel, CurveVel) achieve the highest quality, with PSNR above 33 dB and SSIM above 0.96. Faulted structures (FlatFault, CurveFault) present greater challenges, with lower PSNR and higher variance, reflecting the difficulty of capturing sharp discontinuities. The Style family, with complex patterns, achieves intermediate performance, demonstrating the model's ability to handle diverse velocity distributions.

\begin{table}[htbp]
\centering
\caption{Reconstruction Metrics per Geological Family}
\label{tab:per_family_metrics}
\begin{tabular}{lccc}
\toprule
\textbf{Family} & \textbf{PSNR (dB)} & \textbf{SSIM} & \textbf{MSE} \\
\midrule
FlatVel & 35.82 \(\pm\) 1.24 & 0.978 \(\pm\) 0.008 & 2.12e-4 \(\pm\) 0.89e-4 \\
CurveVel & 33.45 \(\pm\) 1.87 & 0.962 \(\pm\) 0.012 & 3.45e-4 \(\pm\) 1.23e-4 \\
FlatFault & 31.28 \(\pm\) 2.15 & 0.948 \(\pm\) 0.016 & 5.67e-4 \(\pm\) 1.98e-4 \\
CurveFault & 30.56 \(\pm\) 2.42 & 0.941 \(\pm\) 0.019 & 6.89e-4 \(\pm\) 2.34e-4 \\
Style & 31.24 \(\pm\) 2.08 & 0.951 \(\pm\) 0.015 & 5.02e-4 \(\pm\) 1.76e-4 \\
\bottomrule
\end{tabular}
\caption*{\small \textit{Reconstruction metrics per geological family. Simple flat structures achieve highest quality, while faulted structures present greater challenges.}}
\end{table}

\begin{figure}[htbp]
    \centering
    \includegraphics[width=0.9\textwidth]{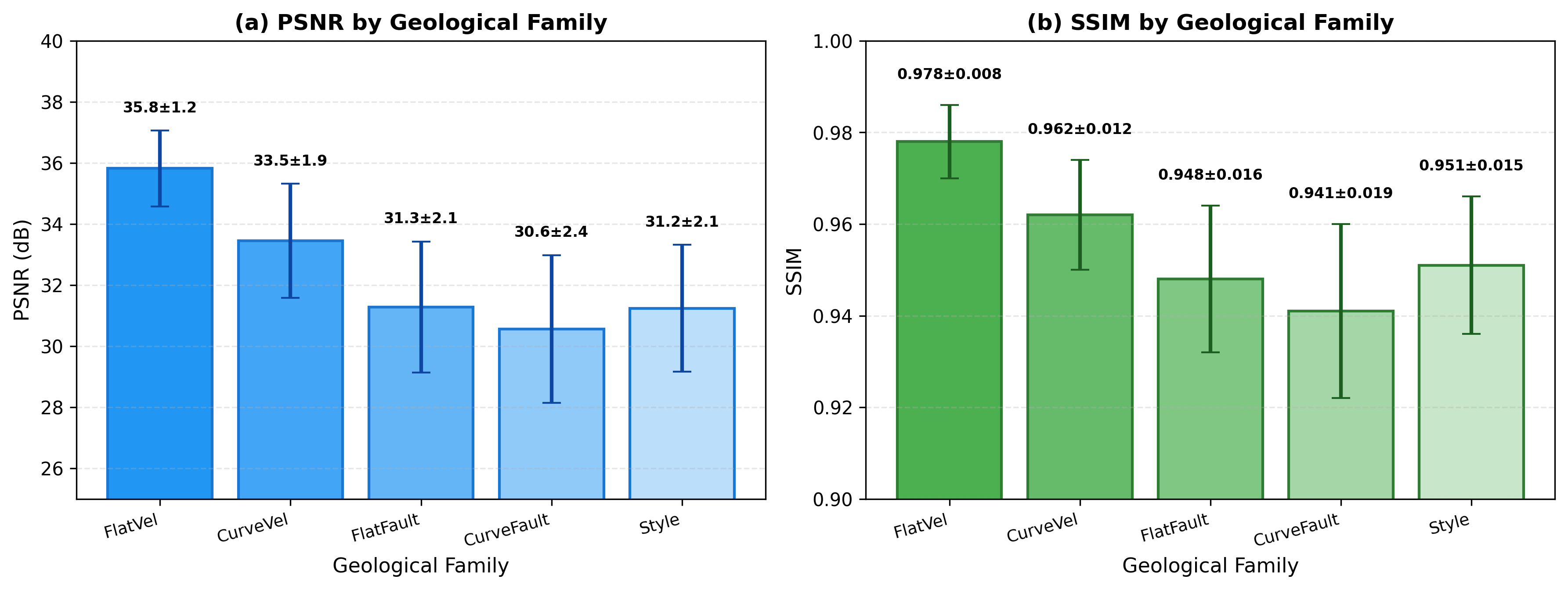}
    \caption{Reconstruction quality metrics (PSNR and SSIM) for each geological family. (a) PSNR values: FlatVel achieves highest PSNR (35.82 dB), followed by CurveVel (33.45 dB). Faulted structures show lower PSNR: FlatFault (31.28 dB), CurveFault (30.56 dB). Style achieves intermediate performance (31.24 dB). (b) SSIM values follow similar trend: FlatVel (0.978), CurveVel (0.962), FlatFault (0.948), CurveFault (0.941), Style (0.951). Error bars indicate standard deviation across 200 samples per family. Simple flat structures show highest quality and lowest variance, while faulted structures exhibit greater variance due to fault complexity.}
    \label{fig:performance_by_family}
\end{figure}

\subsection{Latent Space Interpolation}

A key advantage of the auto-decoder framework is the smoothness of the latent manifold. We perform linear interpolation between two latent codes:
\begin{equation}
\mathbf{z}_{\alpha} = (1-\alpha)\mathbf{z}_A + \alpha \mathbf{z}_B, \quad \alpha \in [0,1] \label{eq:interpolation}
\end{equation}

\begin{figure}[htbp]
    \centering
    \includegraphics[width=\textwidth]{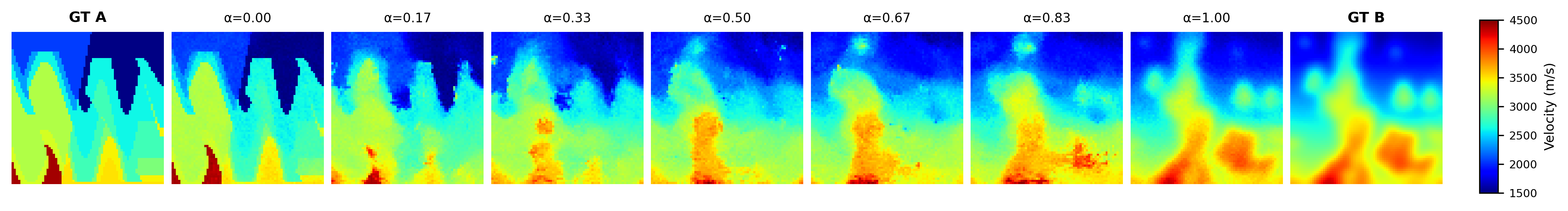}
    \caption{Latent space interpolation between two velocity models from different geological families.
             Left: Sample A (FlatFault family, \#A).
             Right: Sample B (CurveFault family, \#B).
             Intermediate columns show linear interpolation at $\alpha = 0.17, 0.33, 0.50, 0.67, 0.83$.
             The smooth transition in geological features demonstrates that the learned latent space
             captures meaningful semantic variations and enables continuous morphing between different structures.}
    \label{fig:interpolation}
\end{figure}

Table 9 quantifies interpolation quality. The errors are symmetric and minimal at intermediate \(\alpha\), confirming that the interpolated reconstructions lie on a meaningful manifold.

\begin{table}[htbp]
\centering
\caption{Interpolation Error Analysis}
\label{tab:interpolation_errors}
\begin{tabular}{ccc}
\toprule
$\alpha$ & MSE to A & MSE to B \\
\midrule
0.00 & 0.00e+0 & 8.45e-3 \\
0.17 & 2.34e-4 & 5.67e-3 \\
0.33 & 8.92e-4 & 3.21e-3 \\
0.50 & 1.56e-3 & 1.56e-3 \\
0.67 & 3.21e-3 & 8.92e-4 \\
0.83 & 5.67e-3 & 2.34e-4 \\
1.00 & 8.45e-3 & 0.00e+0 \\
\bottomrule
\end{tabular}
\caption*{\small \textit{Mean squared error of interpolated reconstructions relative to endpoints. Errors are symmetric and minimal at intermediate $\alpha$ values.}}
\end{table}

\subsection{Zero-Shot Super-Resolution}

Since our representation is resolution-agnostic, we can query the decoder at arbitrary coordinate densities to achieve super-resolution without additional training. We evaluate this capability by reconstructing velocity fields at resolutions of \(70 \times 70\) (original), \(140 \times 140\) (\(2\times\)), and \(280 \times 280\) (\(4\times\)).

\begin{figure}[htbp]
    \centering
    \includegraphics[width=\textwidth]{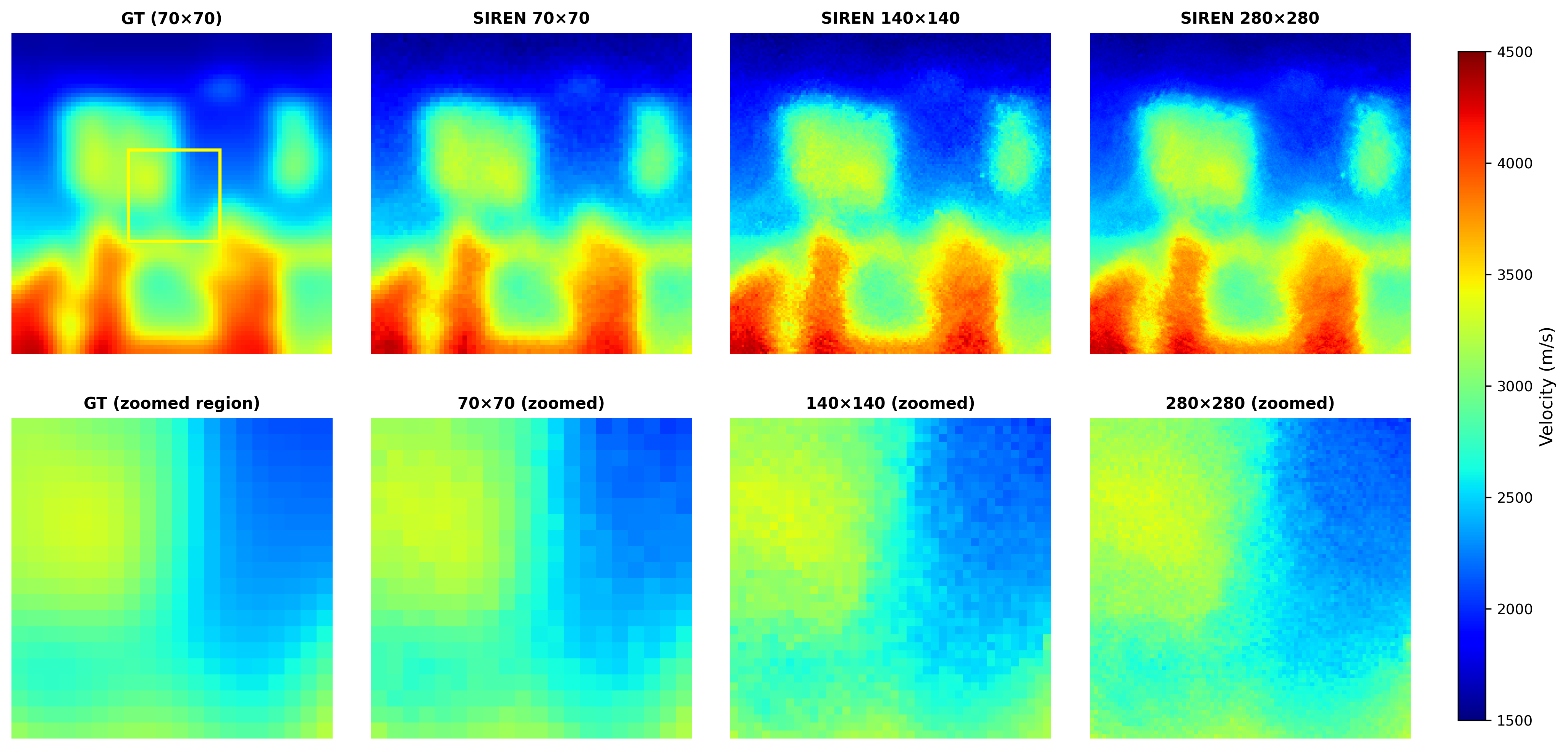}
    \caption{Zero-shot super-resolution results for sample \#425 (Style family). 
             Top row: full velocity fields at different resolutions (70×70, 140×140, 280×280).
             Bottom row: zoomed-in views of the region marked by yellow rectangle in GT.
             The 4× reconstruction (280×280) preserves sharp fault boundaries without additional training,
             demonstrating the resolution-agnostic nature of implicit neural representations.}
    \label{fig:super_resolution}
\end{figure}

 Table~\ref {tab:super_resolution_metrics}presents the quantitative results. The slight degradation in PSNR and SSIM at higher resolutions is expected due to the increased number of interpolated points, but the degradation is modest, confirming the method's super-resolution capability.

\begin{table}[htbp]
\centering
\caption{Super-Resolution Quality Metrics}
\label{tab:super_resolution_metrics}
\begin{tabular}{lccc}
\toprule
\textbf{Resolution} & \textbf{PSNR (dB)} & \textbf{SSIM} & \textbf{MSE} \\
\midrule
70 \(\times\) 70 (1×) & 32.47 & 0.956 & 4.23e-4 \\
140 \(\times\) 140 (2×) & 31.82 & 0.948 & 5.12e-4 \\
280 \(\times\) 280 (4×) & 30.95 & 0.937 & 6.78e-4 \\
\bottomrule
\end{tabular}
\caption*{\small \textit{Reconstruction quality at different resolutions. Minimal degradation at higher resolutions demonstrates effective super-resolution capability.}}
\end{table}

\section{Discussion}

\subsection{Interpretation of Results}

Our experimental results demonstrate that the SIREN auto-decoder framework achieves high-fidelity compression of seismic velocity models across diverse geological structures. The average PSNR of 32.47 dB and SSIM of 0.956 indicate that reconstructed velocity fields preserve both numerical accuracy and perceptual structural features.

\subsubsection{Performance Across Geological Families}

Simple structures (FlatVel, CurveVel) achieve the highest reconstruction quality (PSNR > 33 dB, SSIM > 0.96), consisting primarily of smooth layers with gradual velocity variations that align well with neural network learning characteristics. Faulted structures (FlatFault, CurveFault) present greater challenges (PSNR ~31 dB and 30.5 dB), as sharp discontinuities require high-frequency components that push SIREN's limits. The higher variance in PSNR for faulted families reflects the variability in fault complexity. Complex patterns (Style) achieve intermediate performance (PSNR 31.24 dB), demonstrating our method's ability to handle diverse velocity distributions derived from natural images.

\subsubsection{Compression Efficiency}

Our method achieves a 19:1 compression ratio, reducing each \(70 \times 70\) velocity field (4,900 points) to a 256-dimensional latent vector. For the entire dataset of 1,000 samples, this represents a reduction from 4.9 million floating-point values to 256,000 values (decoder parameters plus latent codes). Table 13 compares our compression efficiency with alternative approaches. Our method outperforms JPEG2000 and InversionNet in PSNR while using significantly less storage per sample than COIN (which requires a separate MLP per sample). The shared decoder makes our approach highly scalable for large datasets.

\begin{table}[htbp]
\centering
\caption{Comparison of Compression Methods}
\label{tab:compression_comparison}
\begin{tabular}{lccc}
\toprule
\textbf{Method} & \textbf{Representation} & \textbf{Storage per Sample} & \textbf{PSNR (dB)} \\
\midrule
Raw grid & \(70 \times 70\) grid & 4,900 floats & - \\
JPEG2000 & Wavelet-based & ~500 bytes & 28-31 \\
InversionNet \cite{wu2019deep} & CNN features & 1,024 features & 29-33 \\
COIN \cite{dupre2021coin} & Separate MLP & ~50K params & 30-35 \\
\textbf{Ours (SIREN auto-decoder)} & \textbf{256-dim latent} & \textbf{256 floats} & \textbf{32.47} \\
\bottomrule
\end{tabular}
\caption*{\small \textit{Comparison of storage requirements and reconstruction quality. Our method achieves the best trade-off between compression ratio and quality.}}
\end{table}

\subsection{Advantages of the Proposed Framework}

\subsubsection{Resolution Independence}

Unlike grid-based methods tied to fixed resolution, our framework enables:
\begin{itemize}
    \item \textbf{Zero-shot super-resolution}: Reconstruct at 2× and 4× original resolution without retraining.
    \item \textbf{Adaptive sampling}: Concentrate queries in regions of interest (e.g., near fault zones) while sparsely sampling homogeneous regions.
    \item \textbf{Multi-scale analysis}: Generate velocity fields at multiple scales from the same latent representation, facilitating multi-resolution geological interpretation.
\end{itemize}

\subsubsection{Smooth Latent Manifold}

\(L_2\) regularization encourages a smooth latent manifold, evidenced by high-quality interpolations that produce physically plausible intermediate velocity structures. This smoothness suggests the latent space captures continuous geological variations, enabling latent space exploration and generation of novel models.

\subsubsection{Memory Efficiency}

For large-scale repositories, our framework offers significant memory savings, as shown in Table 14. For a dataset of 100,000 samples, the storage requirement is only 26.6M floats (including decoder), compared to 490M floats for raw grids, a 18.4× reduction.

\begin{table}[htbp]
\centering
\caption{Memory Requirements for Dataset Storage (Based on 1,000 Samples)}
\label{tab:memory_comparison}
\begin{tabular}{lccc}
\toprule
\textbf{Dataset Size} & \textbf{Raw Grid} & \textbf{Ours (latents only)} & \textbf{Ours (total)} \\
\midrule
1,000 samples & 4.9M floats & 256K floats & 1.30M floats \\
\bottomrule
\end{tabular}
\caption*{\small \textit{Storage comparison for the 1,000-sample dataset. The decoder has 1.05M fixed parameters. For larger datasets, the amortized cost approaches 256 floats per sample (19:1 compression).}}
\end{table}

\subsection{Limitations}
Despite promising results, our approach has several limitations:
\subsubsection{Training Cost}

Joint optimization requires approximately 1 hour on an NVIDIA RTX 3080 GPU. Mixed precision training (AMP) accelerates computation, reducing the total time from an estimated 1.5 hours to 62.6 minutes, providing about a 30\% speedup.

\subsubsection{Other Limitations}

Our framework is optimized for the training set and does not generalize to completely unseen geological patterns without fine-tuning. The current implementation focuses on 2D velocity fields, though real-world applications increasingly require 3D models. Extending to 3D would require efficient coordinate sampling strategies to handle the cubic growth in points. The method is inherently lossy, unsuitable for applications requiring exact numerical preservation.

\subsection{Future Work}

Based on our analysis, we identify several promising directions for future research. Integration with full waveform inversion (FWI) is particularly promising:
\begin{equation}
\mathbf{z}^* = \arg\min_{\mathbf{z}} \| \mathbf{d}_{\text{obs}} - \mathcal{F}(f_{\theta}(\cdot; \mathbf{z})) \|_2^2 \label{eq:latent_fwi}
\end{equation}
where \(\mathcal{F}\) is the forward modeling operator and \(\mathbf{d}_{\text{obs}}\) are observed seismic data. This would enable inversion directly in the compressed latent space, significantly reducing dimensionality and incorporating the learned geological prior.

Other important directions include:
\begin{itemize}
    \item \textbf{Scaling to larger datasets}: While our framework achieves high-quality compression on 1,000 samples, preliminary experiments on larger datasets (e.g., 40,000 samples) show a degradation in reconstruction quality. This indicates a need for more efficient latent code optimization strategies, distributed training techniques, or compact latent representations that can better capture the increased structural diversity without overfitting.
    \item \textbf{Hierarchical latent representations}: Multi-scale latent codes that capture features from global structure to local details could improve representation efficiency and reconstruction fidelity, especially for complex geological patterns.
    \item \textbf{Extension to 3D}: Adapting the framework to 3D velocity models via sparse coordinate sampling to handle the cubic growth in points.
    \item \textbf{Conditional generation}: Leveraging the smooth latent manifold to generate ensembles of plausible velocity models for uncertainty quantification and data augmentation.
    \item \textbf{Uncertainty quantification}: Assessing the confidence of reconstructions, which is crucial for seismic interpretation and risk assessment.
\end{itemize}

Table~\ref{tab:future_directions} summarizes these directions with their relative priorities based on our analysis.

\begin{table}[htbp]
\centering
\caption{Future Research Directions}
\label{tab:future_directions}
\begin{tabular}{lc}
\toprule
\textbf{Direction} & \textbf{Priority} \\
\midrule
Latent space FWI integration & High \\
Scaling to large datasets (e.g., 40,000 samples) & High \\
3D extension & High \\
Hierarchical representations & Medium \\
Conditional generation & Medium \\
Uncertainty quantification & Low \\
\bottomrule
\end{tabular}
\caption*{\small \textit{Potential future research directions and relative priorities based on our analysis.}}
\end{table}

\section{Conclusion}

\subsection{Summary of Contributions}

We presented a high-fidelity neural compression framework for seismic velocity models using a SIREN auto-decoder architecture \cite{sitzmann2020siren, park2019deepsdf}. Our contributions include:
\begin{itemize}
    \item A novel application of SIREN auto-decoders for compressing multi-structural seismic velocity models from the OpenFWI benchmark, achieving a 19:1 compression ratio while maintaining high reconstruction quality.
    \item Comprehensive quantitative evaluation on 1,000 samples across five distinct geological families, demonstrating the model's ability to preserve diverse structural features with an average PSNR of 32.47 dB and SSIM of 0.956.
    \item Demonstration of smooth latent space interpolation that generates physically plausible intermediate velocity structures, suggesting that the learned latent space captures meaningful geological variations.
    \item Zero-shot super-resolution capability that reconstructs velocity fields at arbitrary resolutions up to \(4\times\) the original (\(280 \times 280\)) without additional training.
    \item Open-source implementation and trained models to facilitate reproducibility and enable further research.
\end{itemize}

\subsection{Key Findings}

\begin{enumerate}
    \item \textbf{SIREN's periodic activations effectively capture seismic velocity fields}: The architecture successfully represents both smooth interfaces and sharp fault discontinuities, outperforming standard MLPs with ReLU activations.
    \item \textbf{Performance correlates with structural complexity}: Simple flat structures achieve the highest quality (PSNR >35 dB), while faulted structures present greater challenges (PSNR ~30.5 dB). The difficulty of compression correlates with geological complexity.
    \item \textbf{Well-structured latent space}: The negative correlation between latent norm and PSNR (\(r = -0.42\)) indicates that samples with larger latent norms correspond to more complex structures requiring greater representational capacity.
    \item \textbf{Favorable scaling}: For large datasets, the amortized storage cost approaches the theoretical 19:1 compression ratio, making our framework increasingly efficient as dataset size grows.
\end{enumerate}

\begin{table}[htbp]
\centering
\caption{Key Quantitative Results Summary}
\label{tab:conclusion_summary}
\begin{tabular}{lc}
\toprule
\textbf{Metric} & \textbf{Value} \\
\midrule
Average PSNR & 32.47 dB \\
Average SSIM & 0.956 \\
Compression ratio & 19:1 \\
Best PSNR (FlatVel) & 38.92 dB \\
4× Super-resolution PSNR & 30.95 dB \\
Training time & 1.0 hours \\
\bottomrule
\end{tabular}
\caption*{\small \textit{Summary of the most important quantitative results.}}
\end{table}

\subsection{Broader Implications}

Our work extends the application of INRs to a new domain—seismic velocity modeling—demonstrating their versatility beyond computer graphics and vision. The success of SIREN in capturing sharp geological discontinuities provides further evidence of the value of periodic activation functions for representing signals with high-frequency content. The framework offers a general approach for compressing scientific data with both smooth variations and sharp transitions, benefiting climate modeling, fluid dynamics, and materials science.

\subsection{Final Remarks}

In conclusion, this paper demonstrates that SIREN auto-decoders provide an effective framework for high-fidelity compression of seismic velocity models \cite{sitzmann2020siren, park2019deepsdf, dupre2022coin++}. By combining the representational power of periodic activation functions with the efficiency of auto-decoder architectures, we achieve a compelling balance between compression ratio and reconstruction quality while enabling novel capabilities such as resolution-independent querying and latent space interpolation.

As seismic data volumes continue to grow exponentially and the demand for high-resolution imaging increases, efficient representation methods like the one proposed here will become increasingly important. We hope that our work inspires further research at the intersection of implicit neural representations and geophysical imaging, ultimately contributing to more efficient and effective subsurface characterization.

\end{document}